\documentclass[journal]{IEEEtran}
\usepackage{comment}
\usepackage{graphics}
\usepackage{graphicx}
\usepackage{subcaption}
\usepackage{amsmath} 
\usepackage{amsfonts}
\usepackage{dsfont}
\usepackage{xcolor}
\usepackage[utf8]{inputenc}
\usepackage{booktabs}
\usepackage{colortbl}
\usepackage{xcolor}
\usepackage{pifont}
\usepackage{caption}
\usepackage{url} 
\usepackage{hyperref}

\newcommand{\cmark}{\textcolor{green}{\ding{51}}} 
\newcommand{\xmark}{\textcolor{red}{\ding{55}}} 

\ifCLASSINFOpdf
\else
\fi

\begin{document}

\title{\huge Towards Efficient Multi-LLM Inference: Characterization and Analysis of LLM Routing and Hierarchical Techniques}

\author{Adarsh Prasad Behera, Jaya Prakash Champati, Roberto Morabito, Sasu Tarkoma, and James Gross}

\markboth{Journal of \LaTeX\ Class Files,~Vol.~14, No.~8, August~2015}%
{Shell \MakeLowercase{\textit{et al.}}: Bare Demo of IEEEtran.cls for IEEE Journals}

\maketitle

\begin{abstract}
Recent progress in Language Models (LMs) has dramatically advanced the field of natural language processing (NLP), excelling at tasks like text generation, summarization, and question answering. However, their inference remains computationally expensive and energy-intensive, especially in settings with limited hardware, power, or bandwidth. This makes it difficult to deploy LMs in mobile, edge, or cost-sensitive environments. To address these challenges, recent approaches have introduced multi-LLM intelligent model selection strategies that dynamically allocate computational resources based on query complexity—using lightweight models for simpler queries and escalating to larger models only when necessary. 
This survey explores two complementary strategies for efficient LLM inference: (i) routing, which selects the most suitable model based on the query, and (ii) cascading or hierarchical inference (HI), which escalates queries through a sequence of models until a confident response is found. Both approaches aim to reduce computation by using lightweight models for simpler tasks while offloading only when needed. We provide a comparative analysis of these techniques across key performance metrics, discuss benchmarking efforts, and outline open challenges. Finally, we outline future research directions to enable faster response times, adaptive model selection based on task complexity, and scalable deployment across heterogeneous environments, making LLM-based systems more efficient and accessible for real-world applications.
\end{abstract}



\IEEEpeerreviewmaketitle

\section{Introduction}
 
\IEEEPARstart{L}{arge} Language Models (LLMs), such as Bidirectional Encoder Representations from Transformers (BERT) \cite{devlin2018bert}, Generative Pre-trained Transformer (GPT) \cite{dale2021gpt}, and DeepSeek \cite{bi2024deepseek}, have advanced machine understanding in Natural Language Processing (NLP) by achieving state-of-the-art performance in tasks such as text generation, question answering, and summarization. However, deploying these models at a large scale presents significant computational and financial challenges, particularly for small-to-medium-sized organizations and academic researchers who rely on pre-trained open-source models. While training LLMs demands substantial resources, inference costs remain a persistent concern, especially when high accuracy and responsiveness are required. Addressing these challenges necessitates strategies that optimize computational efficiency while maintaining performance.

Efficient LLM inference is particularly crucial for mobile and edge computing scenarios \cite{qu2024mobile}. Currently, queries from mobile users are predominantly processed by large-scale cloud-based LLMs, including GPT, Claude, Gemini, and Llama. However, cloud-based inference incurs substantial costs due to the reliance on expensive Graphics Processing Units (GPUs) or Tensor Processing Units (TPUs) and potential fees for API calls, limiting scalability and accessibility \cite{chavan2024faster, xu2024survey, kim2025efficient}. An alternative is to shift inference closer to the user by utilizing edge servers equipped with open-source Small Language Models (SLMs) such as Llama 3.2 7B/11B \cite{zheng2024llamafactory}, Phi-3 \cite{brogly2025evaluation}, and Mixtral 7B \cite{doremus2025harnessing}. While this approach reduces latency and reliance on cloud infrastructure, it often compromises response quality \cite{popov2025overview}. Routing \cite{nguyen2024metallm} and Hierarchical Inference (HI) \cite{zhang2023ecoassistant} techniques address this trade-off by dynamically selecting the appropriate model based on task complexity, ensuring a balance between efficiency and performance. 
These techniques prioritize efficient resource allocation by leveraging smaller, cost-effective models for initial processing and offloading to larger models only when necessary. Such adaptive strategies significantly reduce computational overhead while ensuring accuracy \cite{chen2023frugalgpt}. In recent years, advances in these techniques have demonstrated their potential for optimizing LLM deployment, highlighting the need for a comprehensive analysis of their capabilities and future research directions \cite{srivatsa2024harnessing}.

Routing mechanisms assign queries to the most suitable model out of multiple available models based on query complexities and model performances. For example, ZOOTER \cite{lu2023routing} employs reward-based metrics to assign queries to the most suitable models, optimizing both accuracy and cost. HI, on the other hand, employs a cascading \footnote{The terms HI and cascading have been used interchangeably in this paper.} structure where all the queries are processed by lightweight models first, and complex queries will escalate to more powerful models only when additional processing is required. For instance, EcoAssistant \cite{zhang2023ecoassistant} initially utilizes cost-effective models like GPT-3.5-turbo and escalates to GPT-4 when the response generated by the smaller model is not adequate. So in principle, these techniques can be defined as follows:

\textbf{Routing:} Routing systems dynamically assign queries to the most suitable model based on task complexity, accuracy requirements, and latency constraints. This approach ensures that tasks are handled by models that provide the best balance of cost and performance. 

\textbf{Cascading or HI:} Cascading frameworks, on the other hand, prioritize smaller, cost-efficient models for simpler tasks and escalate to larger models only when necessary. This hierarchical inference process minimizes resource usage while maintaining output quality. 

These adaptive strategies facilitate scalable, cost-effective inference, making them advantageous for real-time applications, resource-constrained environments, and high-demand NLP systems \cite{huang2025routereval, hu2024routerbench}.

\begin{figure*}[t]
  \centering
  \subcaptionbox{Efficient routing in edge LLM inference}[.48\linewidth][c]{%
    \includegraphics[width=0.48\textwidth]{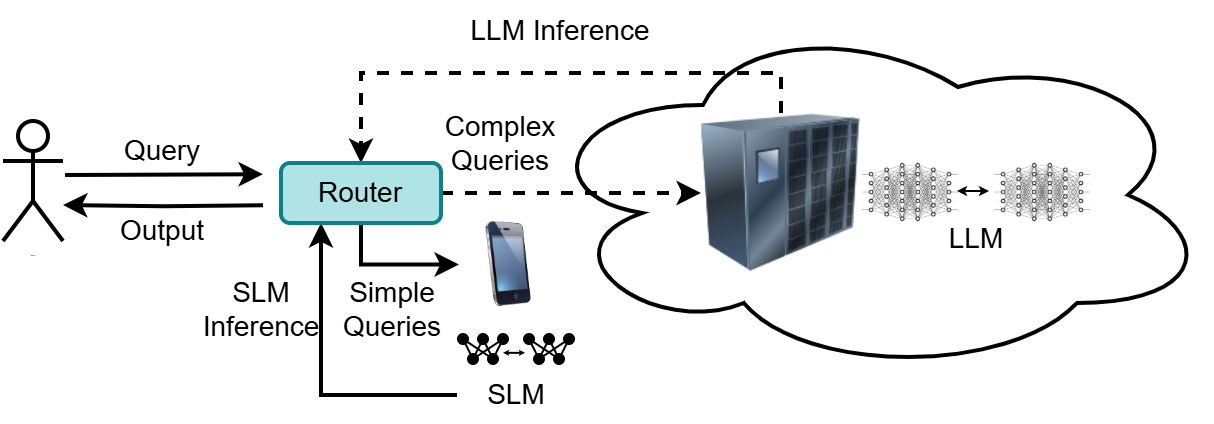}}\quad
  \subcaptionbox{Edge LLM inference using HI}[.48\linewidth][c]{%
     \includegraphics[width=0.48\textwidth]{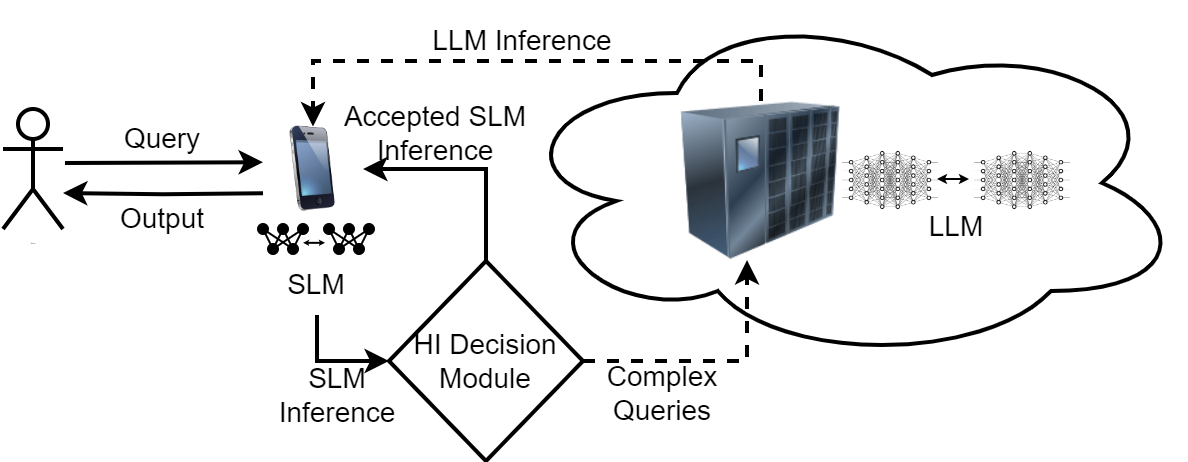}}\quad
    \caption{Efficient LLM inference using different Routing and HI approaches}
    \label{fig:Fig7}
\end{figure*}

\subsection{Related Works}
Efficient inference for LLMs has been a prominent area of research, with numerous surveys examining various techniques to improve performance, reduce costs, and enhance scalability. Several works focus on resource-efficient LLMs by exploring core methods such as pruning, quantization, and distillation. Studies in \cite{xu2024survey, zhou2024survey} highlighted these methods as foundational strategies for minimizing computational overhead while maintaining accuracy. These techniques reduce the size and complexity of LLMs, making them more accessible for deployment in real-world scenarios. Similarly, \cite{chavan2024faster} extended this discussion by addressing the challenges of designing lightweight architectures and accelerating inference. The authors in \cite{bai2024beyond} provided a broader view, examining efficiency strategies across the entire LLM lifecycle, from training to deployment.

Energy efficiency is another critical aspect explored in the literature. Some works \cite{stojkovic2024towards,qu2024mobile} emphasized the importance of developing environmentally conscious LLM deployment strategies and discussed methods for reducing the energy footprint of LLM inference. These works emphasized hardware-aware optimizations, distributed inference techniques, and sustainable AI practices, particularly for edge and mobile deployments. For example, Mobile Edge Intelligence focuses on adapting LLMs for limited-resource environments, such as IoT and mobile systems, where computational and energy constraints are significant. 

At the system level, studies like \cite{li2024llm} and \cite{miao2023towards} explored the infrastructure and algorithms required to efficiently deploy LLMs in real-world settings. These works addressed challenges such as scalable serving frameworks, load balancing, latency optimization, and reliability in large-scale deployments. LLM Inference Serving \cite{li2024llm} emphasized the need for end-to-end systems that can manage high query volumes while meeting performance and cost objectives. In contrast, \cite{miao2023towards} focused on generative tasks, analyzing the integration of serving frameworks with models designed for tasks like summarization and text generation.

Some of the studies such as \cite{yuan2024llm,ding2023efficiency} focused on LLM performance analysis by understanding efficiency trade-offs in LLM inference. These works use frameworks such as roofline models to evaluate computational performance, identify bottlenecks, and highlight opportunities for optimization. These studies are valuable for understanding the trade-offs between computational cost, latency, and model performance in various scenarios. 
Although these surveys provide comprehensive insights into compression, energy efficiency, and infrastructure optimization, they lack a focused exploration of routing and cascading techniques, which are key strategies for adaptive inference in resource-constrained environments.

Recent surveys on LLM routing and model selection have laid important groundwork but leave several critical gaps that our work aims to address. For instance, Chen et al.~\cite{chen2024harnessingensemble} present a broad taxonomy of ensemble methods, including routing as a subset of ensemble-before-inference strategies. However, their focus remains largely architectural, with limited attention to real-world constraints such as latency, energy, and scalability—key concerns in practical deployments. Similarly, Srivatsa et al.~\cite{srivatsa2024harnessing} explore routing for reasoning tasks using policy-based and clustering methods, but restrict their evaluation to academic benchmarks and do not consider cascading or deployment-aware trade-offs. An extended survey by the authors of~\cite{doingmorewithless2024} discusses implementation details for routing systems but lacks a formal framework for comparing methods across cost, performance, and model interoperability.
Additionally, the recent work by Chen et al.~\cite{chen2025survey} surveys collaborative mechanisms between large and small language models, focusing on how they can jointly serve inference tasks through coordination, distillation, and task partitioning. While this work presents useful insights into collaborative architectures, it stops short of providing a unified treatment of routing and HI techniques under deployment constraints. It also lacks a comparative analysis of benchmarks or quantitative models for cost and performance trade-offs, which are essential for real-world applicability.

Unlike prior surveys that emphasize ensemble architectures or accuracy-driven model selection, our survey focuses explicitly on routing and HI as optimization strategies for multi-LLM systems deployed under real-world constraints. We bridge the gap between theoretical methods and practical deployment by analyzing techniques through the lens of compute, memory, energy, latency, financial cost, scalability, and modality compatibility. Additionally, we bring coherence to a fragmented evaluation landscape by comparing key benchmarks—MixInstruct, ROUTERBENCH, and RouterEval—and highlight the need for unified performance metrics in multi-LLM routing, a topic largely overlooked in existing literature.
Our main contributions are fourfold: \textit{(i)} a deployment-aware taxonomy of routing and HI techniques categorized by the resource constraints they address; \textit{(ii)} a comparative review of major routing benchmarks and the introduction of a unified evaluation metric, \textit{Inference Efficiency Score (IES)}, to assess cost-performance trade-offs; \textit{(iii)} a synthesis of existing methods with practical deployment scenarios such as edge inference and mobile assistants; and \textit{(iv)} a forward-looking analysis of emerging research directions, including multimodal routing, adaptive inference, and integration with reasoning-capable LLMs. Together, these contributions offer a foundation for designing scalable, efficient, and context-aware LLM routing systems.


One notable exception to these techniques is the Mixture of Experts (MoE) \cite{cai2024survey} approach, which was recently employed by the DeepSeek model~\cite{bi2024deepseek}. MoE selectively activates the most relevant expert sub-networks based on input characteristics, thereby minimizing unnecessary computations and improving efficiency. However, it differs fundamentally from the routing and cascading techniques discussed in this survey, as MoE operates within a single monolithic model architecture. In contrast, routing and HI approaches function across a collection of distinct models, typically involving combinations of SLMs and LLMs. Since the focus of this survey is on multi-LLM setups used to optimize inference efficiency across diverse deployment environments, MoE falls outside the scope of this work and is not considered in further detail.

The remainder of the paper is structured as follows. In Section~\ref{sec2}, we provide the necessary background for this study, including an overview of LLMs, SLMs, and the various resource constraints involved. Section~\ref{sec3} presents the system model along with a detailed discussion of routing and hierarchical inference (HI) techniques. Evaluation methodologies and performance metrics are discussed in Section~\ref{sec4}, followed by challenges and potential future directions in Section~\ref{sec5}. Finally, we conclude the paper in Section~\ref{sec6}.


\section{Background} \label{sec2}
\subsection{Overview of LLM architecture and inference process}
\subsubsection{Transformers} LLMs are predominantly constructed using transformer-based architectures. The introduction of transformers \cite{vaswani2017attention} has revolutionized NLP, achieving remarkable results in a wide range of language tasks, including text classification \cite{soyalp2021improving}, machine translation \cite{ott2018scaling}, and question-answering \cite{guan2022block}. A notable example is BERT \cite{devlin2018bert}, which has set new benchmarks in question-answering tasks by efficiently capturing contextual information. 

\begin{figure}[htbp]
\centerline{\includegraphics[width=3.5in]{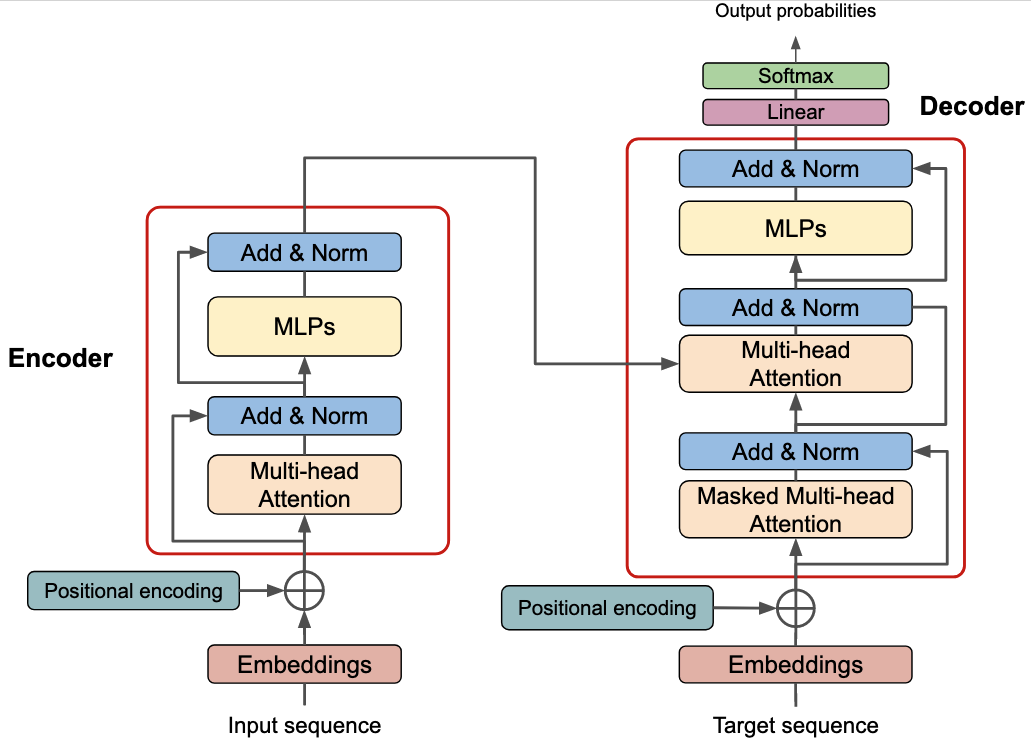}}
\caption{Transformer Architecture.}
\label{bertarch}
\end{figure}

Transformers employ self-attention mechanisms to capture complex dependencies between elements in a sequence and effectively model long-range relationships. The architecture is based on an encoder-decoder framework, where each component consists of stacked layers incorporating multi-head self-attention mechanisms, feed-forward networks (FFNs), and layer normalization. The encoder processes input sequences into rich contextual representations, while the decoder uses these representations, alongside previously generated tokens, to generate output sequences. 
Given an input sequence of tokens, we represent it as a matrix of embeddings \( \mathbf{X} \in \mathbb{R}^{n \times d} \), where \( n \) denotes the sequence length and \( d \) is the dimensionality of each token embedding. In the self-attention mechanism, the model computes attention scores using the scaled dot-product attention formula:

\[
\text{Attention}(\mathbf{Q}, \mathbf{K}, \mathbf{V}) = \text{softmax}\left( \frac{\mathbf{Q} \mathbf{K}^\top}{\sqrt{d_k}} \right) \mathbf{V}.
\]
where, \( \mathbf{Q}, \mathbf{K}, \mathbf{V} \in \mathbb{R}^{n \times d_k} \) are the query, key, and value matrices, respectively. These are obtained by linearly projecting the input matrix \( \mathbf{X} \) using trainable weight matrices \( \mathbf{W}_Q, \mathbf{W}_K, \mathbf{W}_V \in \mathbb{R}^{d \times d_k} \) as follows: \( \mathbf{Q} = \mathbf{X}\mathbf{W}_Q \), \( \mathbf{K} = \mathbf{X}\mathbf{W}_K \), and \( \mathbf{V} = \mathbf{X}\mathbf{W}_V \). The product \( \mathbf{Q}\mathbf{K}^\top \in \mathbb{R}^{n \times n} \) computes pairwise similarity scores between tokens via dot product. These scores are scaled by \( \sqrt{d_k} \) to prevent the softmax function from entering regions with very small gradients, which stabilizes training. The softmax operation then normalizes these scores row-wise to produce attention weights, which are used to compute a weighted sum over the value vectors in \( \mathbf{V} \). The result is a context-aware representation of the input sequence that captures dependencies between tokens.
 The multi-head attention (MHA) mechanism extends this by performing multiple attention calculations in parallel:
\[
\text{MHA}(\mathbf{Q}, \mathbf{K}, \mathbf{V}) = \text{Concat}(\text{head}_1, \dots, \text{head}_h)\mathbf{W}_O,
\]
where each head is computed as \( \text{head}_i = \text{Attention}(\mathbf{Q}_i, \mathbf{K}_i, \mathbf{V}_i) \), and \( \mathbf{W}_O \in \mathbb{R}^{hd_k \times d} \) is a learned projection matrix.

Although attention mechanisms have been applied in feed-forward and recurrent networks \cite{chaudhari2021attentive, de2022attention}, transformers rely exclusively on this specialized multi-head attention implementation. This design enables efficient parallelization, making transformers particularly suitable for scaling to high-complexity models and processing large datasets. Alternative approaches, such as hard attention \cite{vinyals2015show}, are inherently stochastic and require Monte Carlo sampling to determine attention positions, which adds computational overhead.

Unlike convolutional or recursive architectures \cite{goodfellow2016deep, lecun2015deep, graves2012long}, transformers require minimal prior knowledge of problem structures. This flexibility makes them ideal for pre-training on large-scale unlabeled datasets through pretext tasks \cite{vaswani2017attention, devlin2018bert}. Such pre-training enables the generation of expressive and generalizable representations, \( \mathbf{H} \), that effectively capture relationships among entities in the data. These representations serve as a robust foundation for supervised fine-tuning in downstream tasks, further enhancing the versatility and impact of transformers.

\subsubsection{LLMs} The scalability of transformers has been a key driver behind the rapid advancement of LLMs. Many LLMs have been developed and refined based on transformer architecture, with leading AI companies actively designing and deploying these models across diverse domains. For instance, OpenAI’s GPT-3 \cite{dale2021gpt} achieves exceptional performance in tasks such as text generation and machine translation, setting benchmarks for chat-based applications. Google’s Med-PaLM \cite{singhal2023towards} demonstrates expertise in the medical domain by providing expert-level diagnoses and guidance. Similarly, Facebook’s DEiT \cite{touvron2021training} incorporates self-supervised learning into the transformer framework to achieve state-of-the-art image classification with limited annotated data. These LLMs are trained on extensive and diverse datasets \cite{lin2024splitlora}.

LLM architectures can be broadly categorized into three types: encoder-only, encoder-decoder, and decoder-only models. Encoder-only LLMs, such as ALBERT \cite{lan2019albert}, consist exclusively of encoder components and leverage advanced transformer-based architectures \cite{vaswani2017attention}. Given an input sequence \( \mathbf{X} \in \mathbb{R}^{n \times d} \), the encoder generates contextualized token representations \( \mathbf{H} = \text{Encoder}(\mathbf{X}) \), where \( \mathbf{H} \in \mathbb{R}^{n \times d_{\text{model}}} \). These models excel in tasks like text classification, sentence similarity, and language understanding, owing to their efficient feature extraction and versatile representations. Despite lacking decoders for output sequence generation, encoder-only LLMs remain highly effective for analyzing and understanding text.

Encoder-decoder LLMs, such as T5 \cite{raffel2020exploring}, combine encoder and decoder components to perform sequence-to-sequence tasks. The encoder processes input sequences \( \mathbf{X} \) into contextual representations \( \mathbf{H}_{\text{enc}} \), while the decoder uses these representations along with previously generated tokens \( \mathbf{Y}_{<t} \) to produce output sequences \( \mathbf{Y} = \{y_1, y_2, \dots, y_T\} \). This process can be expressed as:
\[
\mathbf{H}_{\text{enc}} = \text{Encoder}(\mathbf{X}), \quad y_t = \text{Decoder}(\mathbf{H}_{\text{enc}}, \mathbf{Y}_{<t}),
\]
where \( t \) is the current timestep. Encoder-decoder models excel in applications such as machine translation, text summarization, and question answering, where capturing intricate linguistic structures and contextual dependencies is essential.

Decoder-only LLMs, such as the GPT series \cite{dale2021gpt, sanderson2023gpt}, represent a prominent class of language models designed specifically for generative tasks. These models utilize an autoregressive decoding approach, in which tokens are generated sequentially, each conditioned on the previously generated tokens. The probability of generating a sequence \( \mathbf{Y} = \{y_1, y_2, \dots, y_T\} \) is modeled as:

\[
P(\mathbf{Y}) = \prod_{t=1}^T P(y_t \mid \mathbf{Y}_{<t}),
\]

where \( \mathbf{Y}_{<t} = \{y_1, y_2, \dots, y_{t-1}\} \) denotes all tokens preceding the current token \( y_t \). This autoregressive formulation enables the model to produce coherent and contextually relevant outputs, making it particularly effective for tasks such as language generation, text completion, and dialogue response generation. By leveraging large-scale training datasets and the inherent scalability of transformer architectures, these three LLM categories collectively demonstrate exceptional versatility and performance across a wide range of domains and applications.

\textbf{Multimodal LLMs:}
Traditional LLMs, such as GPT-3 and BERT \cite{brown2020language, devlin2018bert}, are designed for unimodal textual data, limiting their usefulness in real-world scenarios that involve multiple modalities, such as images, audio, or sensor data \cite{ngiam2011multimodal, baltruvsaitis2018multimodal, tsai2019multimodal}. To address this, multimodal LLMs have emerged, integrating diverse data types within a unified framework. For instance, GPT-4 \cite{openai2023gpt4} can process both text and images, supporting tasks like image captioning and aiding speech recognition. These models leverage cross-modal fusion and interactive learning to handle complex inputs \cite{liu2021cross}. By incorporating foundation models trained on different modalities—such as vision and audio—multimodal LLMs extend the capabilities of traditional language models \cite{radford2021learning, bommasani2021opportunities}. However, integrating these models requires effective cross-modal alignment, typically achieved through multimodal pre-training and instruction tuning.

Multimodal pre-training learns shared representations across different modalities by optimizing objectives on multimodal datasets such as XText \cite{xtext}. Given two modalities \( \mathbf{M}_1 \) and \( \mathbf{M}_2 \), the model constructs a joint representation \( \mathbf{H} \):

\[
\mathbf{H} = f_{\text{align}}(\mathbf{M}_1, \mathbf{M}_2),
\]
where \( f_{\text{align}} \) is a neural alignment function. This alignment process is optimized using contrastive loss:

\[
\mathcal{L}_{\text{contrastive}} = -\log \frac{\exp(\text{sim}(\mathbf{H}_{\mathbf{M}_1}, \mathbf{H}_{\mathbf{M}_2}) / \tau)}{\sum_{j} \exp(\text{sim}(\mathbf{H}_{\mathbf{M}_1}, \mathbf{H}_{\mathbf{M}_{2,j}}) / \tau)},
\]
where \( \text{sim} \) is a similarity function (e.g., cosine similarity), and \( \tau \) is a temperature parameter \cite{li2021align}. This alignment strengthens inter-modal relationships, improving cross-modal task performance.

Instruction-tuning, on the other hand, refines pre-trained models for task-specific objectives by optimizing labeled multimodal data. For a given task \( \mathcal{T} \) with input-output pairs \( (\mathbf{M}_\text{input}, \mathbf{M}_\text{output}) \) and $N$ training examples, the model minimizes the loss as:

\[
\mathcal{L}_{\text{task}} = \sum_{i=1}^N \mathcal{L}(\hat{\mathbf{M}}_\text{output}^i, \mathbf{M}_\text{output}^i),
\]
where \( \hat{\mathbf{M}}_\text{output} \) is the model's prediction, and \( \mathcal{L} \) is a task-specific loss function, such as cross-entropy loss \cite{ouyang2022training}. Instruction-tuning enhances the model’s ability to generalize to unseen instructions, improving zero-shot learning.


\subsection{Different Resource Constraints}  
Performing LLM inference in a device-edge-cloud setting presents several resource constraints due to the size and complexity of LLMs and the limitations of deployment environments. Broadly, these constraints can be categorized into general hardware and operational limitations, followed by LLM-specific challenges. Among the general constraints, \textbf{compute constraints} arise because edge devices often lack powerful processing units such as GPUs, TPUs, or dedicated AI accelerators, making efficient execution of large models difficult. \textbf{Memory constraints} further exacerbate this issue, as LLMs require substantial memory for storing model weights, intermediate computations, and input-output data, which may exceed the capabilities of resource-limited devices. Additionally, \textbf{energy constraints} play a crucial role in mobile and IoT applications, where battery-powered devices have limited power supplies, and LLM inference can drain batteries rapidly due to intensive computational and memory demands. Moreover, \textbf{latency constraints} are particularly problematic for real-time applications that require fast responses, as on-device inference may introduce computational delays, and cloud-based inference, while alleviating local processing burdens, incurs additional transmission delays that can hinder real-time usability. 

Beyond these general constraints, there are specific challenges associated with deploying LLMs effectively. \textbf{Financial constraints} arise due to the high cost of deploying specialized hardware such as GPUs or TPUs on edge devices, while cloud-based solutions impose recurring costs for compute time, storage, and data transfer, making large-scale deployment expensive. Another key challenge is \textbf{scalability constraints}, where applications serving many users, such as large-scale chatbots or recommendation systems, must balance computational load, latency, and accuracy within constrained environments. Finally, \textbf{modality constraints} limit LLMs, as they primarily rely on text inputs and struggle to process multimodal data such as images or audio effectively. Although multimodal LLMs aim to bridge this gap, challenges in aligning representations and managing cross-modal dependencies persist, necessitating advancements in robust multimodal techniques. Addressing these constraints is critical for optimizing LLM deployment across diverse environments.

\subsection{Small Language Models (SLMs)}
SLMs \cite{dong2024hymba} have emerged as a viable alternative to LLMs, offering a balance between efficiency and performance. Popular SLMs such as Llama 3.2 7B/11B, Phi-3, Mixtral 7B, and Mistral 7B have been developed to provide high-quality language understanding and generation while requiring significantly fewer computational resources than their larger counterparts \cite{wang2024comprehensive}. The motivation behind SLM research stems from the growing demand for models that can operate on resource-constrained devices, such as smartphones, edge servers, and embedded systems, without relying on expensive cloud-based inference. By leveraging innovations in training efficiency, architecture optimization, and knowledge distillation, SLMs achieve competitive performance on common NLP tasks while being more accessible and cost-effective for developers and businesses \cite{wang2024comprehensive}.

Despite their advantages, SLMs face notable limitations. Due to their reduced parameter count, they often struggle with tasks requiring deep reasoning, long-term memory retention, or highly nuanced contextual understanding compared to larger models like GPT-4, Gemini 1.5, or Claude 3 \cite{lu2024small}. Additionally, SLMs may exhibit hallucinations, biases, or factual inconsistencies, as their training data is more condensed \cite{xu2024hallucination, huang2024survey}. While techniques such as fine-tuning, retrieval-augmented generation (RAG), and prompt engineering help mitigate these issues \cite{lin2024towards}, they cannot fully replace the broader knowledge and flexibility of LLMs. Nevertheless, ongoing research into model compression, adaptive scaling, and hybrid inference strategies continues to enhance the capabilities of SLMs, making them an increasingly practical choice for real-world applications where cost, latency, and deployment constraints are primary concerns \cite{abstreiter2025sometimes}.

\section{LLM Inference using Routing and HI Techniques}\label{sec3}
In this section, we first present the system model from a cost-centric perspective, detailing how various deployment constraints impact inference decisions. We then review a range of algorithms that employ routing or HI strategies to enable efficient multi-LLM inference under these constraints. A summarized comparison of the key characteristics of these techniques is provided in Table~\ref{tab:routing_hi_summary}.

\subsection{System Model}  
As discussed previously, LLM inference is constrained by computational cost, memory, energy cost, latency, financial feasibility, scalability, and modality compatibility. Let \( \mathcal{M} = \{ M_1, M_2, \dots, M_K \} \) denote a set of LLMs, where \( M_1 \) is the least resource-intensive model and \( M_K \) is the most resource-intensive. Given an input query \( q \), the routing function \( R(q) \) determines the most suitable model \( M_k \) based on a learned function \( f_R(q, \theta) \):
\[
R(q) = M_k \quad \text{where} \quad k = f_R(q, \theta),
\]
where $\theta$ denotes the parameters of the learned routing function. The routing decision must satisfy multiple constraints to optimize efficiency while maintaining accuracy. Let \( C(M_k) \) denote the total cost of selecting \( M_k \), defined as:
\begin{align*}
    C(M_k) = C_{\text{compute}}(M_k) + C_{\text{memory}}(M_k) + C_{\text{energy}}(M_k) + \\C_{\text{latency}}(M_k) +
    C_{\text{financial}}(M_k) + \\
    C_{\text{scalability}}(M_k) + C_{\text{modality}}(M_k).
\end{align*}

Each cost component is modeled as follows:

\begin{itemize}
    \item \textbf{Computational Cost:} Defined as the FLOPs required for inference:
    \[
    C_{\text{compute}}(M_k) = \beta_k \cdot \text{FLOPs}(M_k),
    \]
    where $\text{FLOPs}(M_k)$ is the number of floating-point operations required for inference, and $\beta_k$ is the computational efficiency factor.
    
    \item \textbf{Memory Cost:} Defined as the memory footprint of the model:
    \[
    C_{\text{memory}}(M_k) = \gamma_k \cdot \text{Mem}(M_k),
    \]
    where $\text{Mem}(M_k)$ is the memory footprint of model $M_k$, and $\gamma_k$ is a weight related to memory bandwidth limitations.
    
    \item \textbf{Energy Cost:} Defined as the energy consumption per inference:
    \[
    C_{\text{energy}}(M_k) = \delta_k \cdot P_k \cdot T_k,
    \]
    where $P_k$ is the power consumption and $T_k$ is the inference time of model $M_k$, and $\delta_k$ is an energy scaling coefficient.
    
    \item \textbf{Latency Cost:} Defined as the inference time penalty:
    \[
    C_{\text{latency}}(M_k) = \lambda_k \cdot T_k,
    \]
    where $\lambda_k$ reflects application-specific latency sensitivity, and $T_k$ is the response time of $M_k$.
    
    \item \textbf{Financial Cost:} Defined as the monetary cost of running inference:
    \[
    C_{\text{financial}}(M_k) = \mu_k \cdot \text{API\_Cost}(M_k),
    \]
    where \( \mu_k \) accounts for cloud API costs.
    
    \item \textbf{Scalability Cost:} Defined as the overhead incurred in handling increased query loads:
    \[
    C_{\text{scalability}}(M_k) = \rho_k \cdot \text{Load}(M_k),
    \]
    where \( \rho_k \) is the scaling factor and \(\text{Load}(M_k)\) represents number of requests the model can handle.
    
    \item \textbf{Modality Cost:} Defined as the penalty when a model is incompatible with the required input modalities:
    \[
    C_{\text{modality}}(M_k) = \sigma_k \cdot \mathds{1}_{M_k \notin \mathcal{M}_{\text{compatible}}},
    \]
    where \( \mathds{1}_{M_k \notin \mathcal{M}_{\text{compatible}}} \) is an indicator function that is 1 if \( M_k \) does not support the input modality.
\end{itemize}

The HI framework applies cascading selection based on confidence scores \( s_k(q) \), where inference escalates to a larger model only if the model $k$'s confidence is below threshold \( \tau_k \):
\[
M(q) =
\begin{cases} 
M_1(q), & s_1(q) \geq \tau_1 \\  
M_2(q), & s_1(q) < \tau_1, s_2(q) \geq \tau_2 \\  
\vdots & \\  
M_K(q), & s_{K-1}(q) < \tau_{K-1}.
\end{cases}
\]
The overall objective is to minimize cost while maintaining accuracy:
\[
\mathcal{L} = \mathbb{E}_{q \sim \mathcal{D}} \bigg[ \ell(M_k(q), M^*(q)) + \sum_{k=1}^{K} \mathds{1}_{s_k(q) < \tau_k} C(M_k) \bigg].
\]
where $\mathcal{D}$ is the query distribution, $\ell(\cdot, \cdot)$ is a task-specific loss function (e.g., cross-entropy), and $M^*(q)$ is the ground-truth or ideal model output for query $q$.
This formulation integrates multiple real-world constraints, ensuring that LLM inference remains computationally efficient, memory-aware, energy-efficient, financially viable, latency-optimized, scalable, and modality-compatible.

We now turn to practical implementations of routing and HI strategies that, to some extent, aim to operationalize the efficiency trade-offs described in our cost model. These methods differ in how they estimate query complexity, assign models, and manage resource constraints in real-time. For clarity, we organize them into two broad categories: routing-based techniques and HI approaches, each reflecting distinct model selection dynamics.

\subsection{Routing-Based Techniques}
\subsubsection{Tryage}
In \cite{hari2023tryage}, the authors proposed a routing algorithm Tryage, that selects optimal models for user prompts by predicting downstream model performance and integrating user-defined constraints. The router analyzes prompts dynamically using a Q-learning-inspired approach, where a predictive model estimates the performance of each expert model in a library. The routing decision minimizes a loss function that combines predicted accuracy with weighted user constraints, such as model size, recency, and latency. To achieve this, the algorithm uses supervised learning to train the router to approximate model-specific losses. Once trained, the router dynamically allocates prompts to the most suitable model.

\subsubsection{ZOOTER} Another routing method that efficiently assigns queries to the most suitable LLMs based on their expertise is proposed in \cite{lu2023routing}. The ZOOTER algorithm uses reward distillation, where off-the-shelf reward models provide scalar rewards to train a routing function. This function learns to predict the likelihood that a specific model is optimal for a given query. During inference, the routing function directly assigns queries to the most capable model, avoiding the computational overhead of generating outputs from all candidates. Additionally, tag-based label enhancement is applied to reduce noise in reward signals, improving the routing function’s robustness and accuracy.

\subsubsection{FORC} A meta model driven approach for efficient routing is proposed in \cite{vsakota2024fly}. FORC (Fly-Swat or Cannon) uses a lightweight meta-model trained on different datasets to predict the performance and cost of each LLM for a given query before invoking any model. Each query is assigned to a single LLM based on predicted cost-performance trade-offs, without the need to sequentially test multiple LLMs. The meta-model considers multiple user-defined constraints, such as inference budgets and accuracy requirements. Its design ensures that the framework is generalizable and does not require retraining for specific datasets.

\subsubsection{Routoo} Routoo \cite{mohammadshahi2024leeroo} targets efficient query routing in scenarios with thousands of open-source LLMs. It uses an LLM as a Performance Predictor, to score potential model performance for a query. A cost-aware selector then maximizes performance within budget constraints. This approach extends earlier frameworks by handling a vast pool of models, emphasizing scalability in open-source contexts.

\subsubsection{HybridLLM} Ding et al. \cite{ding2024hybrid} introduced a hybrid routing approach using a SLM (Llama 2) and a larger LLM (GPT-3.5 turbo). They trained a BERT-style encoder using BARTscore to assess query quality and directed it to the appropriate model. This work integrates adaptive routing with quality-aware strategies, complementing earlier studies by focusing on maintaining high-quality outputs while reducing computational cost.

\subsubsection{OptLLM} the authors in \cite{liu2024optllm} formulated query routing as a multi-objective optimization problem, balancing performance and cost. They trained random forest models to predict LLM accuracy on specific queries based on pre-collected responses. By modeling trade-offs between performance and computational cost, this approach allows for fine-grained query assignment to LLMs. The focus on predictive modeling aligns with the dynamic and adaptive frameworks in the earlier works but emphasizes cost-performance trade-offs through robust optimization.

\subsubsection{MetaLLM} A dynamic framework designed to optimize the process of routing queries to the most appropriate LLM for classification tasks is proposed in \cite{nguyen2024metallm}. In this framework, the selection of the optimal LLM is treated as a multi-armed bandit (MAB) problem, where each LLM is considered as an ``arm" with associated performance and cost values. MetaLLM learns to balance these two factors to determine the best LLM for each query without requiring an exhaustive search through all options. The system dynamically routes queries to a pool of available LLMs, including models like OpenAI’s GPT \cite{chatgpt}, Amazon's Titan \cite{titan}, Anthropic's Claude \cite{claude}, and Meta's LLaMa \cite{llama}, based on their performance and cost characteristics. It significantly improves accuracy and reduces costs by up to 50\% to 70\% compared to using a single model. MetaLLM's approach allows for effective zero-shot classification, optimizing query performance while minimizing the overall cost, a crucial consideration for large-scale applications.

\subsubsection{RouteLLM} In \cite{ong2024routellm}, Ong et al. introduced a novel approach to efficiently route queries among multiple LLMs (read SLMs and LLMs) based on learned preferences to optimize computational cost and performance. The authors propose a framework where routing decisions are guided by a model trained on preference data, which reflects user preferences for outputs generated by different LLMs in various scenarios. This preference data is used to train a routing policy, enabling the system to dynamically select the most appropriate LLM for a given input. A standout method, Similarity-Weighted Ranking, computes query similarity to known responses during inference without training. This method enables dynamic routing decisions, offloading to the LLM only when the SLM inference will be insufficient, reducing overall computation. This paper lays a foundation for preference-aware query routing, emphasizing minimal additional training overheads.

\begin{table*}[htbp]
\centering
\caption{Summary of Routing and Hierarchical Inference Techniques}
\label{tab:routing_hi_summary}
\begin{tabular}{@{}cccc@{}}
\toprule
\textbf{Technique (Routing/HI)} & \textbf{Supervision Type} & \textbf{Routing/Cascading Strategy} \\
\midrule
\textbf{Tryage} \cite{hari2023tryage} & Supervised & Q-learning inspired performance prediction \\

\textbf{ZOOTER} \cite{lu2023routing} & Supervised (via reward distillation) & Reward-guided routing using learned function \\

\textbf{FORC} \cite{vsakota2024fly} & Supervised & Meta-model predicts cost-performance trade-off \\

\textbf{Routoo} \cite{mohammadshahi2024leeroo} & Supervised & LLM-based performance predictor with cost-aware selector \\

\textbf{HybridLLM} \cite{ding2024hybrid} & Supervised & BERT-style encoder with quality estimation (BARTScore) \\

\textbf{OptLLM} \cite{liu2024optllm} & Supervised & Random forest predictor for model accuracy and cost  \\

\textbf{MetaLLM} \cite{nguyen2024metallm} & Unsupervised (Multi-Armed Bandit) & MAB with reward balancing cost and performance \\

\textbf{RouteLLM} \cite{ong2024routellm} & Supervised / Rule-based (Similarity Matching) & Preference-based policy and similarity-weighted ranking \\

\hline

\textbf{FrugalGPT} \cite{chen2023frugalgpt} & Rule-based / Heuristic & Cascading with generation scoring + thresholds \\

\textbf{EcoAssistant} \cite{zhang2023ecoassistant} & Supervised / Feedback-driven & Escalation based on user feedback and execution checks \\

\textbf{Cache \& Distil} \cite{ramirez2023cache} & Supervised (Student-Teacher) & Confidence-based escalation using uncertainty metrics \\

\textbf{Automix} \cite{aggarwal2023automix} & Unsupervised / POMDP-based & Verifier + POMDP-guided routing \\

\textbf{Efficient Hybrid Decoding} \cite{ms2024efficient} & Supervised (via reward model) & Token-level scoring with reward threshold \\

\textbf{Uncertainty-Based Selection} \cite{ramirez2024optimising} & Unsupervised / Online Learning & Marginal sampling with dynamic threshold \\
\bottomrule
\end{tabular}
\end{table*}

\subsection{Hierarchical Inference Approaches}
\subsubsection{FrugalGPT} The first hierarchical approach called FrugalGPT was proposed by Chen et al. \cite{chen2023frugalgpt}. This approach sequences queries through a list of LLMs ranked by cost and reliability. Starting with inexpensive models, it offloads the query to more expensive models like GPT-4 only if earlier responses fail a reliability threshold. This decision is guided by a generation scoring function that evaluates the quality of each model's response and an LLM router that determines the optimal order of querying based on user-defined budget and accuracy constraints. In addition to cascading, FrugalGPT employs complementary strategies like prompt adaptation, which reduces input size to cut costs, and LLM approximation, which uses cached responses or fine-tuned smaller models to avoid redundant API calls. For example, query concatenation processes multiple queries simultaneously, minimizing prompt overhead. These techniques enable up to 98\% cost savings while achieving comparable or even superior accuracy compared to using only high-end LLMs like GPT-4.

\subsubsection{EcoAssistant} In \cite{zhang2023ecoassistant}, a hierarchical approach to optimize cost-efficiency is proposed. The system starts with the most cost-effective LLM (e.g., GPT-3.5-turbo) to address a query. If this model fails to resolve the issue, the system escalates to more powerful models like GPT-4. Query resolution success is determined by two methods: user feedback and automated checks. After a model generates a response, user feedback is gathered to assess if the query was resolved successfully. Additionally, if the query involves code execution (such as API calls), the system checks if the code runs correctly and produces the expected results. This tiered approach minimizes the use of expensive models, reducing costs while still ensuring high performance. The solution demonstration technique further enhances accuracy by storing successfully resolved query-code pairs. When new queries arise, the system retrieves and appends relevant past solutions to improve response quality. Together, these techniques allow EcoAssistant to resolve queries efficiently, surpassing GPT-4’s success rate by 10\% while reducing costs by over 50\%.

\subsubsection{Cache \& Distil} Cache \& Distil \cite{ramirez2023cache} uses a smaller, local student model (SLM) trained continuously on responses from a more expensive teacher model (LLM). Queries are first routed to the SLM, which attempts to provide a prediction. If the SLM confidence is low, the query is escalated to the LLM for accurate resolution. This HI system employs active learning techniques like margin sampling (queries with the smallest margin between top predictions) and prediction entropy (queries with high output uncertainty) to decide when to consult the LLM. Additionally, query by committee \cite{seung1992query} measures disagreement among multiple student models, while coreset sampling prioritizes diverse queries for annotation. These methods ensure the SLM improves over time while minimizing offloading to the LLM.

\subsubsection{Automix} In \cite{aggarwal2023automix}, an HI system is proposed, where an SLM evaluates its own outputs using a verifier prompt designed by the authors. Given the noisy nature of such verification, the system employs a Partially Observable Markov Decision Process (POMDP) framework to guide routing decisions. This adaptive approach ensures robust selection among multiple LLMs, optimizing efficiency and performance in dynamic language modeling scenarios. 

\subsubsection{Efficient Hybrid Decoding} In \cite{ms2024efficient}, the authors introduce a hybrid decoding framework that combines an on-device SLM with a larger, cloud-based LLM. In this setup, the SLM generates tokens, which are then evaluated by a reward model trained on synthetic data reflecting outputs from both the SLM and the cloud LLM. The reward model scores each token based on its expected alignment with the estimated token distribution of the cloud LLM . If the score exceeds a threshold, the token is accepted; otherwise, the task is offloaded, and the cloud LLM is queried to generate the next tokens.

\subsubsection{Uncertainty-Based Two-Tier Selection}
Another hierarchical framework to optimize resource usage when working with LLMs is proposed in \cite{ramirez2024optimising}. The SLM handles initial predictions, while the decision to offload to a larger LLM is taken based on marginal sampling. Marginal sampling captures the uncertainty of the SLM's predictions by measuring the probability difference between the top two token predictions. Using a dynamic threshold, which is updated through online learning, the method decides whether offloading is necessary. Unlike other frameworks, this approach does not require an auxiliary model for decision-making. 





\section{Evaluation and Metrics}\label{sec4}

As discussed in Section \ref{sec3}, the effectiveness of routing and HI strategies depends not only on model accuracy but also on system-level trade-offs such as latency, energy use, and cost. These constraints shape model selection decisions. In this section, we review existing evaluation methodologies and benchmarks that attempt to quantify these trade-offs. However, current evaluation efforts remain fragmented, with few benchmarks capturing the full complexity of multi-LLM inference under real-world constraints.

\subsection{Qualitative assessment for routing and HI approaches}

\begin{table*}[htbp]
\centering
\caption{Comparison of techniques addressing various constraints}
\begin{tabular}{@{}cccccccccc@{}}
\toprule
\textbf{Technique} & \textbf{Compute} & \textbf{Memory} & \textbf{Energy} & \textbf{Latency} & \textbf{Financial} & \textbf{Scalability} & \textbf{Modality} \\ \midrule
Tryage \cite{hari2023tryage}         & \cmark & \xmark & \xmark & \cmark & \cmark & \xmark & \xmark \\
ZOOTER \cite{lu2023routing}          & \cmark & \cmark & \xmark & \cmark & \cmark & \cmark & \xmark \\
FORC \cite{vsakota2024fly}           & \cmark & \cmark & \cmark & \cmark & \cmark & \xmark & \xmark \\
Routoo \cite{mohammadshahi2024leeroo}& \cmark & \cmark & \xmark & \cmark & \cmark & \cmark & \xmark \\
HybridLLM \cite{ding2024hybrid}      & \cmark & \cmark & \cmark & \cmark & \cmark & \xmark & \xmark \\
OptLLM \cite{liu2024optllm}          & \cmark & \cmark & \xmark & \cmark & \cmark & \cmark & \xmark \\
MetaLLM \cite{nguyen2024metallm}     & \cmark & \xmark & \xmark & \cmark & \cmark & \cmark & \xmark \\
RouteLLM \cite{ong2024routellm}      & \cmark & \xmark & \cmark & \cmark & \cmark & \cmark & \xmark \\
\hline
FrugalGPT \cite{chen2023frugalgpt}   & \cmark & \cmark & \cmark & \cmark & \cmark & \cmark & \xmark \\
EcoAssistant \cite{zhang2023ecoassistant} & \cmark & \cmark & \cmark & \cmark & \cmark & \xmark & \xmark \\
Cache \& Distil \cite{ramirez2023cache} & \cmark & \cmark & \cmark & \cmark & \cmark & \xmark & \xmark \\
Automix \cite{aggarwal2023automix}   & \cmark & \cmark & \cmark & \cmark & \cmark & \xmark & \xmark \\
Efficient Hybrid Decoding \cite{ms2024efficient} & \cmark & \cmark & \cmark & \cmark & \cmark & \xmark & \xmark \\
Uncertainty-Based Selection \cite{ramirez2024optimising} & \cmark & \cmark & \cmark & \cmark & \cmark & \xmark & \xmark \\ \bottomrule
\end{tabular}
\label{tab:constraints_comparison}
\end{table*}

\begin{table*}[ht]
\centering
\caption{Advantages and Disadvantages of Routing and HI Techniques}
\label{tab:advantages_disadvantages}
\begin{tabular}{@{}|p{4cm}|p{6cm}|p{6cm}|@{}}
\hline
\textbf{Technique} & \textbf{Advantages} & \textbf{Disadvantages} \\
\hline
\textbf{\centering Tryage} & Fast routing for low-latency applications; reduces compute load & No memory or energy optimization; limited scalability \\
\hline
\textbf{ZOOTER} & Strong support for compute, memory, latency, and scalability; reward-based routing & Does not address energy or modality constraints \\
\hline
\textbf{FORC} & Broad constraint coverage including energy; effective for edge environments & Limited scalability; high implementation complexity \\
\hline
\textbf{Routoo} & Latency and cost-efficient routing; scalable to multiple LLMs & No energy or modality considerations \\
\hline
\textbf{HybridLLM} & Efficient across compute, memory, and energy; practical for cost-aware systems & Lacks scalability support; modality-limited \\
\hline
\textbf{OptLLM} & Balances latency and scalability well; cost-efficient deployment & No energy optimization; limited memory strategies \\
\hline
\textbf{MetaLLM} & Cost-focused routing; scalable architecture & No support for memory or energy constraints \\
\hline
\textbf{RouteLLM} & Covers compute, latency, energy, and scalability; edge-aware & Does not optimize memory; modality not addressed \\
\hline
\hline
\textbf{FrugalGPT} & Wide constraint coverage; strong in financial and energy efficiency & May involve complex pipeline tuning; no modality support \\
\hline
\textbf{EcoAssistant} & Good latency and energy efficiency; uses dynamic routing & No scalability or modality support \\
\hline
\textbf{Cache \& Distil} & Caching boosts speed and energy savings; memory-efficient & Static design not suited for scalable deployments \\
\hline
\textbf{Automix} & Effective across memory, latency, and energy; good for mobile use & Not scalable; limited model diversity \\
\hline
\textbf{Efficient Hybrid Decoding} & Balanced across compute, memory, energy, latency & Not scalable; modality and adaptability limitations \\
\hline
\textbf{Uncertainty-Based Selection} & Dynamic model selection improves energy and cost use & Not scalable or modality-aware; memory handling is limited \\
\hline
\end{tabular}
\end{table*}

The core strength of routing and HI techniques lies in reducing computational overhead and latency by dynamically assigning queries to models based on task complexity. Systems employing routing strategies often utilize mechanisms such as confidence scoring, reinforcement learning, or performance-based heuristics to defer simple queries to lightweight models while escalating more complex ones only when necessary. Techniques like \textbf{ZOOTER} and \textbf{Routoo} exemplify this approach by delivering low-latency responses without significant loss in accuracy. Conversely, HI techniques such as \textbf{FORC} and \textbf{FrugalGPT} focus on minimizing economic costs through selective offloading, often avoiding expensive models like GPT-4. These capabilities make routing and HI frameworks particularly suited for latency and cost-sensitive deployments such as mobile assistants and edge-based inference systems \cite{qu2024mobile, zheng2025review, chen2025survey}. A comparative overview of how each method addresses specific resource constraints is provided in Table~\ref{tab:constraints_comparison}, and their key advantages and trade-offs are summarized in Table~\ref{tab:advantages_disadvantages}.

However, attention to memory and energy constraints remains uneven across the literature. HI systems such as \textbf{FORC}, \textbf{HybridLLM}, and \textbf{FrugalGPT} integrate explicit mechanisms to reduce energy usage, such as early exits and runtime pruning. In contrast, several other techniques prioritize compute-time efficiency while largely neglecting energy profiles—an oversight with implications for battery-constrained deployments. Memory handling is similarly inconsistent. \textbf{ZOOTER}, for example, reduces memory usage by loading models selectively, whereas \textbf{Tryage} and \textbf{MetaLLM} rely on static pipelines, limiting their adaptability to low-memory environments. These issues are particularly important for on-device inference, where both memory and power are tightly constrained \cite{xu2024device}.

Scalability and modality awareness are arguably the least explored dimensions. Techniques like \textbf{OptLLM} and \textbf{RouteLLM} introduce features for multi-user scalability or distributed inference, but most frameworks are designed for single-query or batch-style usage. This limits their applicability in high-throughput environments, such as large-scale chatbots or multi-tenant APIs. Modality support is even more limited; none of the reviewed methods are designed to route across multimodal LLMs. As real-world applications increasingly demand models that understand and generate across modalities—text, vision, and speech—this remains a critical gap. Future work in routing and HI should focus on closing these limitations, enabling not just fast and cost-efficient inference, but also scalable and modality-aware systems that adapt robustly to real-world usage demands.

\subsection{Benchmarks for Multi-LLM inference systems}
Despite the growing adoption of routing and HI techniques, standardized benchmarks remain scarce, and current tools often focus on narrow task types or static evaluation conditions. In this sub-section, we briefly discuss these benchmarks for evaluation of routing and HI techniques in multi-LLM systems and a comparative analysis of these benchmarks is presented in Table \ref{tab:routing_benchmarks}.

\textbf{MixInstruct}, introduced by Jiang et al.~\cite{jiang2023llm}, is a benchmark designed to evaluate routing strategies across a broad spectrum of instruction-following tasks. It includes 11 popular open-source LLMs and a diverse set of prompts varying in complexity, domain, and intent. This diversity allows testing whether routing systems can correctly match tasks to appropriate models—using lightweight models for simpler prompts and escalating when needed.

\textbf{ROUTERBENCH}~\cite{hu2024routerbench} provides a large-scale, standardized framework for evaluating routing strategies under real-world constraints. It measures model selection performance across dimensions such as latency, accuracy, cost, and response quality. By simulating routing scenarios and using a weighted scoring function, it enables fine-grained comparisons and policy optimization. Studies using ROUTERBENCH have shown that LLM routing can lead to substantial reductions in API costs and latency in production-scale LLM applications. For instance, IBM Research reported that their routing approach achieved a 5-cent saving per query compared to using GPT-4 alone \cite{ibm2024}. Additionally, experiments with hybrid routing systems have shown cost reductions of up to 75\% while maintaining approximately 90\% of GPT-4's performance \cite{premai2024}.

\textbf{RouterEval}~\cite{huang2025routereval} is a lightweight evaluation framework that provides ground-truth model assignments for each query. This setup allows precise measurement of routing accuracy—i.e., how often the selected model provides the best trade-off between cost and performance. RouterEval is particularly well-suited for validating learned routing functions and heuristic policies in multi-LLM settings.

\begin{table*}[ht]
\centering
\caption{Comparison of LLM Routing Benchmarks}
\label{tab:routing_benchmarks}
\begin{tabular}{|p{2.5cm}|p{3.5cm}|p{5cm}|p{5cm}|}
\hline
\textbf{Features $\downarrow$} & \textbf{MixInstruct ~\cite{jiang2023llm}} & \textbf{RouterBench~\cite{hu2024routerbench}} & \textbf{RouterEval~\cite{huang2025routereval}} \\
\hline
\textbf{Focus} & Instruction-following tasks & Systematic assessment of routing systems & Model-level scaling and routing decision quality \\
\hline
\textbf{Dataset Size} & 11 models, diverse prompts & 405k+ inference outcomes & 200M+ performance records \\
\hline
\textbf{Evaluation Metrics} & BARTScore & Latency, accuracy, cost, response quality & Ground-truth model assignments \\
\hline
\textbf{Routing Emphasis} & Per-query model selection & Routing under various constraints & Optimal model selection based on correctness and efficiency \\
\hline
\textbf{Strengths} & Diverse tasks for instruction-following & Standardized framework for routing evaluation & Precise testbed for routing decision quality\\
\hline
\end{tabular}
\end{table*}


\subsection{The Need of Unified Evaluation Metrics for Multi-LLM Inference}
\label{sec:unified_metrics}

Based on our analyses, existing evaluation efforts predominantly rely on a set of isolated performance indicators, such as accuracy, latency, or monetary cost. While these metrics are individually relevant, they do not fully capture the systemic trade-offs that characterize practical deployments of routing and HI strategies, especially when considering constrained computing environments. As introduced in Section~\ref{sec3}, LLM selection is governed by a composite cost model that accounts for computational, energy, latency, memory, financial, scalability, and modality constraints. However, current benchmarks discussed in this section do not yet provide an integrated view of how these constraints interact during inference or how decision-making policies perform under multiple objectives.

To address this limitation, we argue that a unified metric is needed, one that not only reflects the accuracy of the selected model but also quantifies the cost-effectiveness and responsiveness of the inference process. Inspired by system-level metrics in edge computing environments~\cite{jang2025edge}, we propose an evaluation approach based on a utility-driven formulation. Specifically, we define a \textit{multi-LLM Inference Efficiency Score} (IES), which encapsulates model quality and responsiveness, normalized by the cost function $C(M_k)$ defined earlier:
\begin{equation}
\text{IES}(q) = \frac{\alpha \cdot Q(q) + (1 - \alpha) \cdot R(q)}{C(M_k)},
\label{eq:ies_def}
\end{equation}
where $Q(q)$ denotes a task-specific quality score (e.g., accuracy, BLEU, ROUGE), $R(q)$ denotes a responsiveness measure (e.g., time-to-first-token, escalation depth, or total inference time), and $\alpha \in [0,1]$ is a tunable weight controlling the trade-off between quality and responsiveness.

This formulation naturally integrates with the system cost model introduced in Section~\ref{sec3}, and can be adapted for both routing and HI strategies. For routing techniques, IES can be computed per-query to measure the effectiveness of a single-shot model selection. For HI systems, a cumulative version of IES may be considered, aggregating the score across the sequence of models traversed during escalation.

The adoption of unified metrics such as IES would allow future benchmarks to go beyond point-wise accuracy and latency measurements, enabling fairer and more holistic comparisons between model selection strategies. In addition, such metrics would facilitate the study of adaptive policies that dynamically prioritize speed, cost, or quality based on real-time constraints and application-level goals.

Importantly, the relevance of unified evaluation metrics becomes even more pronounced with the increasing deployment of LLM inference pipelines at the edge~\cite{he2024ultraeval, agrawal2024etalon}. In such contexts, traditional cloud-centric metrics, often optimized around throughput or token-based pricing, fail to capture the impact of resource heterogeneity and platform-specific efficiency. IES, by explicitly normalizing utility against system cost, provides a principled way to evaluate model selection policies in environments with constrained compute and energy profiles, supporting the emerging trend of deploying small- and medium-scale LMs across mobile, embedded and on-premise systems \cite{zheng2025review}.

\section{Challenges and Future Directions}\label{sec5}
While the goal of this paper is to highlight the great promises of multi-LLM inference systems, several open challenges remain before these techniques can be reliably deployed at scale. Beyond the well-studied cost-performance trade-offs, real-world scenarios introduce complex dimensions such as infrastructure heterogeneity, evolving input modalities, privacy constraints, and the need for adaptive behavior under uncertainty. In this section, we discuss critical and emerging challenges, ranging from multimodal integration and distributed coordination to evaluation frameworks and privacy-aware routing, grounded in recent literature. This discussion aims to clearly highlight current limitations, but also to point toward key research directions necessary for building LLM inference frameworks with greater robustness and scalability.

\subsection{Integration of Multimodality}

A key open challenge is extending LLM routing systems beyond text to handle images, audio, and other modalities. Modern LLMs are predominantly text-focused, and effectively fusing multiple input types in a single pipeline remains non-trivial~\cite{zhang2024mm}. Current multimodal LLM architectures often bolt together separate modality-specific encoders (e.g., a vision model alongside a language model), which incurs additional latency and complexity~\cite{yu2025mquant}. This makes real-time multimodal inference difficult, as visual or audio tokens can slow down overall response speed~\cite{yu2025mquant}.
Most routing and HI techniques reviewed in this survey are designed exclusively for text inputs. However, emerging LLM use cases, such as visual question answering or audio-captioning assistants, require processing multiple modalities. Current methods like RouteLLM~\cite{ong2024routellm} or ZOOTER~\cite{lu2023routing} do not consider query modality in their routing logic. This creates a gap when queries include image-text pairs or spoken commands, where the choice of model must depend on both the input type and the model's multimodal capabilities. To better understand the challenges in multimodal routing, one must consider both (i) the structure of the query (single-modality vs. composite) and (ii) the capabilities of the model pool (modality coverage, performance profile). These dimensions introduce a non-trivial matching problem not captured by text-only routing functions.

Moreover, adaptive routing for multimodal inputs is largely unexplored. Deciding which model (vision vs. language) should lead for a given query type is challenging without sophisticated policies. Standardized evaluation benchmarks for multimodal LLM routing are still emerging, somehow delaying rigorous comparison of approaches~\cite{alsaad2024multimodal}. In summary, seamlessly combining text with other data types demands new fusion techniques and coordination strategies. Future research should focus on efficient multimodal fusion mechanisms (e.g., shared latent spaces or cross-modal adapters) and develop benchmarks that capture end-to-end performance on tasks involving mixed modalities~\cite{alsaad2024multimodal}. Tackling these issues is crucial to enable LLM cascades that can, for example, interpret a user’s question while also analyzing an image or audio clip in tandem.

Finally, routing logic must account for both input complexity and input type. For instance, if a lightweight text-only model scores high on a task, but cannot handle an embedded image, the router must fall back to a multimodal-capable LLM like GPT-4V. Yet, few current frameworks operationalize such modality compatibility checks within their routing objectives. As formalized in Section~\ref{sec3}, this kind of mismatch can be modeled through a modality penalty term, $C_{\text{modality}}(M_k)$, representing the cost incurred when a selected model lacks the necessary modality support for a given query. In addition, real-time multimodal inference often leads to elevated costs across multiple system dimensions \cite{he2024ma}: visual or audio inputs typically increase latency ($C_{\text{latency}}(M_k)$), memory footprint ($C_{\text{memory}}(M_k)$), and energy consumption ($C_{\text{energy}}(M_k)$). Despite this, such costs are rarely integrated into routing objectives or used to guide decision-making. Even when multimodal models are available, routing based on cross-modal complexity remains largely unexplored. Unlike textual complexity, where proxies such as perplexity, sequence length, or syntactic structure can inform model selection~\cite{north2023lexical}, cross-modal queries often require reasoning over heterogeneous inputs (e.g., aligning visual context with textual captions or audio cues)~\cite{liu2024collaborative}. Designing lightweight predictors of such complexity, capable of estimating fusion effort or modality-bridging difficulty, remains a promising direction for routing-aware multimodal inference~\cite{baltruvsaitis2018multimodal, hu2024routerbench}.

\subsection{Scalability and Distributed Inference}

Scalability is a fundamental concern as multi-LLM inference systems transition from prototype environments to large-scale deployments. This includes both infrastructure-level scalability (e.g., routing thousands of queries per second) and deployment scalability across heterogeneous environments, including edge and cloud tiers. One major challenge is orchestrating distributed inference across many servers or devices. Splitting a large model across machines (via model or tensor parallelism) introduces heavy communication overhead—frequent synchronization (e.g., all-reduce operations) between nodes can drastically slow throughput~\cite{zhang2025communication}. For instance, coordinating partial results across devices often becomes a bandwidth bottleneck~\cite{zhang2025communication}.
While multi-LLM routing and HI offer computational savings at query-time, their real-world scalability remains limited by two key factors: \emph{(i)} model selection overhead, and \emph{(ii)} system coordination under dynamic loads. In large deployments, such as customer service platforms or content-moderation pipelines, routing decisions must be made in milliseconds and scale across hundreds of requests per second. Still, most existing techniques, such as Tryage~\cite{hari2023tryage} or MetaLLM~\cite{nguyen2024metallm}, assume static resource availability and do not incorporate system-level feedback (e.g., current GPU load or network latency) into routing logic.
A growing body of work has begun exploring feedback-aware routing, where infrastructure-level signals guide inference decisions. Current routing functions typically prioritize model accuracy or query complexity, but often neglect dynamic feedback from the system environment. For instance, signals such as queue length, current GPU load, or network latency are rarely considered in routing policies. However, recent work on system-level schedulers~\cite{miao2023towards, hu2024routerbench} shows that integrating such feedback can significantly improve responsiveness and resource utilization. Efficient scheduling of requests in a cluster is equally critical, as simplistic job allocation strategies can result in GPU under-utilization or long queuing delays. Intelligent cluster-level routing, such as reinforcement learning-based workload schedulers, has been shown to reduce latency by dynamically batching and placing queries based on real-time resource status~\cite{jain2024intelligent}.

Another dimension of scalability is edge-cloud integration. Routing frameworks increasingly consider whether a query can be answered by a smaller edge model versus deferring to a powerful cloud LLM. However, deploying LLMs on edge devices is limited by hardware constraints (compute, memory, energy). Only smaller models fit on-device, causing a performance gap compared to cloud-scale models~\cite{zhang2024llm}. For example, a mobile assistant may respond to general queries locally for speed, but escalate complex requests (e.g., code generation or multi-hop reasoning) to a cloud-based LLM. This leads to a trade-off between local inference speed and the richer capabilities of larger remote models. Future systems must dynamically balance this trade-off such as, for example, quickly predicting if a query is simple enough for an edge model or if it requires escalation to the cloud. Research should also explore elastic scaling policies that can spin up additional model instances on demand and load-balance across them in real time.

HI methods like FrugalGPT~\cite{chen2023frugalgpt} and EcoAssistant~\cite{zhang2023ecoassistant} rely on sequential escalation to larger models, which may introduce latency bottlenecks at scale. For instance, escalating a batch of low-confidence queries to GPT-4 can saturate cloud resources rapidly. Parallelism-aware variants or batch-level early exits could reduce latency under heavy workloads, but this remains an underexplored area.

\begin{table*}[ht]
\centering
\caption{Summary of Open Questions in Multi-LLM Inference Systems}
\label{tab:open_questions}
\begin{tabular}{|p{3.5cm}|p{12cm}|}
\hline
\textbf{Challenge Area} & \textbf{Open Questions} \\
\hline
\textbf{Integrating Multimodality} &
\begin{itemize}
    \item How can routers assess both input-modality structure (e.g., text, image-text, audio) and model capability (modality support) during selection?
    \item Can modality-aware scoring functions be learned efficiently without retraining multimodal models?
    \item What are the trade-offs between early modality fusion (e.g., embedding-level) and late fusion (model-level) in hierarchical pipelines?
\end{itemize} \\
\hline
\textbf{Scalability and Distributed Inference} &
\begin{itemize}
    \item How can routing decisions incorporate runtime system metrics without introducing excessive overhead?
    \item Can reinforcement learning or scheduling theory help design scalable, adaptive inference pipelines across cloud and edge?
    \item Are there optimal batching or caching strategies for routing at scale?
\end{itemize} \\
\hline
\textbf{Evaluation and Debugging} &
\begin{itemize}
    \item What metrics best capture the adaptability and robustness of routing/HI strategies under shifting query loads?
    \item How can evaluation pipelines simulate realistic system dynamics, including latency spikes, model downtime, or usage drift?
    \item Can interactive debugging tools help developers audit and fine-tune routing decisions post-deployment?
\end{itemize} \\
\hline
\textbf{Adaptive Routing Strategies} &
\begin{itemize}
    \item Can routers learn optimal model-selection strategies that generalize across workloads and tasks?
    \item How can online learning and feedback signals be incorporated without interrupting the inference pipeline?
    \item What role can cost-aware or utility-driven learning play in routing optimization?
    \item How can routing functions accurately predict reasoning complexity and tool dependencies before inference?
    \item How should cost models evolve to reflect the true computational and interaction overhead of reasoning?
\end{itemize} \\
\hline
\textbf{Privacy and Security Considerations} &
\begin{itemize}
    \item Can privacy constraints be learned jointly with routing decisions?
    \item How can sensitivity of queries be estimated efficiently at inference time?
    \item What guarantees are possible for privacy-aware inference at the edge?
\end{itemize} \\
\hline
\end{tabular}
\end{table*}

\subsection{Evaluation and Debugging}

With growing complexity in multi-LLM pipelines, robust evaluation and debugging methodologies are crucial to ensure reliability. A core issue is the lack of standardized metrics and benchmarks to assess a routing system’s overall performance. Most existing LLM evaluations target single-model accuracy or latency in isolation, which do not capture the combined efficiency and quality of a cascaded system. The absence of common benchmarks for LLM routing has been noted as a barrier to progress~\cite{hu2024routerbench}. Encouragingly, recent efforts like RouterBench~\cite{hu2024routerbench} have begun to fill this gap by providing systematic evaluation frameworks and large-scale datasets for multi-LLM routing comparisons. Researchers argue for greater standardization in how routing strategies are tested, calling for community benchmarks and consistent metrics~\cite{varangot2025doing}. For adaptive systems such as FrugalGPT~\cite{chen2023frugalgpt} or Cache \& Distil~\cite{ramirez2023cache}, which involve multiple models and decision stages, end-to-end metrics like offloading rate, escalation depth, decision stability, route consistency across similar queries, and cost-normalized quality are crucial. More holistic evaluation frameworks that simulate realistic conditions (e.g., bursty traffic, network contention, or shifting workloads) are still lacking.

Beyond aggregate metrics, interpretable evaluation is an open challenge. Traditional metrics (throughput, accuracy) may not fully reflect whether a router is making good decisions. New metrics are needed to benchmark system-wide efficiency, adaptability, and even fairness of routing decisions~\cite{varangot2025doing}. For instance, fairness-aware metrics could track whether certain query types are disproportionately escalated, and route consistency could measure if minor variations in input lead to unstable decisions.

One overlooked challenge is debugging the decision function itself. When a router chooses the wrong model, it is often unclear whether the issue stems from poor input features, inaccurate scoring, or outdated confidence thresholds. Explainable AI (XAI) techniques could be adapted to expose internal routing logic, by for example highlighting which input features (such as token type, query length, or complexity score) influenced the decision to send a query to Model A vs. B, and guide retraining or threshold adjustment~\cite{bommasani2021opportunities}. This has stimulated interest in applying such techniques to LLM routers, to make their decision logic more interpretable~\cite{cambria2024xai}. Greater interpretability would help developers identify biases or failure modes in the routing policy (e.g., consistently misrouting certain query types). For example, understanding which signals triggered a fallback to a larger model could reveal if the trigger logic is overly sensitive or miscalibrated. In future research, developing automated debugging tools, such as log analyzers that highlight anomalous routing patterns, or visualization methods that trace model selection logic, will be important. Holistic evaluation suites that test not just accuracy and cost, but also the correctness of routing decisions and ease of debugging, are needed to drive the field forward~\cite{cambria2024xai,varangot2025doing}.

\subsection{Adaptive Routing Strategies}

Most multi-LLM inference systems today use relatively static routing policies or simple heuristics (for instance, always trying a cheap model first, then escalating if needed). A major research direction is to make routing adaptive, that is able to intelligently adjust model selection on a per-query basis and improve over time. One challenge is that many existing solutions optimize a fixed cost–accuracy trade-off and are not easily customizable. For example, some cascades always aim to minimize compute cost, which can hurt accuracy, and their routing criteria are hand-tuned for a specific scenario. Studies have observed that current LLM cascades largely focus on cost reduction, with any accuracy boosts often coming from non-generalizable tweaks or requiring all candidate models to run (ensemble-style) for each query~\cite{sikeridis2024pickllm}. This static approach misses opportunities to optimize other objectives or to learn from experience. Emerging work is casting model selection as a learning problem: e.g. using reinforcement learning or bandit algorithms to learn the optimal routing policy. A recent framework called PickLLM demonstrates this idea by training a router that weighs query-specific factors like expected accuracy, latency, and API cost, and decides which model to use for each query on the fly~\cite{sikeridis2024pickllm}. Over time, such a router can adapt to usage patterns, deferring to larger models only when absolutely necessary. Early results show that RL-assisted routing can substantially improve efficiency metrics (like reducing cost per query and latency) without sacrificing accuracy~\cite{sikeridis2024pickllm}. Another aspect of adaptivity is handling non-stationary conditions: as user queries evolve (in topic or difficulty), or as new models become available, the routing strategy should update itself. Online learning for routers is largely unexplored, but would enable systems that self-tune in production. The future of multi-LLM inference will likely involve self-optimizing routers that continuously balance multiple objectives (accuracy, latency, cost, etc.) and personalize decisions to context. Developing lightweight learning algorithms that can run alongside inference (without large overhead) is an open research problem. In summary, making routing policies more dynamic and learning-driven stands as a promising direction to boost overall system performance and adaptability.

\subsubsection*{Routing for Reasoning-Capable Models}

Another promising direction for future research is extending routing and HI frameworks to support advanced reasoning models that incorporate tool use and dynamic function calls. Unlike standard LLMs, reasoning-capable models—such as those used in OpenAI’s \texttt{o3}/\texttt{o4} systems—operate at a finer granularity, performing token-by-token inference interleaved with runtime decisions like tool invocation or function execution. These capabilities introduce a new layer of complexity: inference becomes procedural and conditional, rather than a single-pass generation process.

In such settings, routing becomes a more nuanced task. Rather than relying solely on input complexity or modality, routing functions must now account for \emph{reasoning depth} and \emph{tool dependencies}. For instance, simple factual queries might be efficiently handled by small models, whereas queries that trigger external tool-calls (e.g., calculator, database, or API lookups) should be routed to models equipped for procedural reasoning. This requires routing functions that can predict not only the difficulty of a task but also the structural requirements of the reasoning process, including the likelihood and type of tool use.

Moreover, the traditional cost models employed in routing systems must be adapted. Beyond standard metrics like FLOPs, memory, and latency, we now need to factor in \emph{reasoning complexity}—which can include the number of intermediate steps, types of tools invoked, and the branching logic required during execution. This suggests a finer decomposition of computational cost: into token-level inference and tool-level interaction overhead. Such decomposition is especially vital for latency-sensitive or budget-constrained deployments, where reasoning behavior can substantially impact response time and cost. It enables the possibility of orchestrating general-purpose and specialized models more intelligently, balancing performance with resource efficiency. Developing adaptive routing policies that can anticipate reasoning demands and tool usage will be a crucial step toward the next generation of efficient, modular LLM systems.

\subsection{Privacy and Security Considerations}

In multi-LLM deployments, especially those spanning cloud and edge tiers, user data privacy and system security are increasingly critical. Routing a query to an external model (e.g., a third-party API) can inadvertently expose personally identifiable information (PII) unless privacy constraints are explicitly modeled. While many existing cascades optimize for performance and cost~\cite{zhang2024llm}, they often ignore privacy as a routing objective. In practice, applications in healthcare, finance, or personal assistance may require that queries containing sensitive data be handled only by local or secure models, even at the expense of reduced accuracy.

Privacy-aware routing presents multiple challenges. First, systems must detect or annotate the privacy sensitivity of each query, e.g., via client-side classifiers~\cite{albanese2023text} or metadata tagging. Second, the router must respect data-locality policies: for instance, ensuring that queries flagged as containing PII are never offloaded to the cloud. Third, the system may leverage privacy-preserving inference mechanisms such as trusted execution environments (TEEs)~\cite{karanjai2024trusted}, homomorphic encryption~\cite{rho2024encryption}, or anonymization pipelines~\cite{wiest2024anonymizing}, but these techniques introduce added complexity and latency that must be reflected in the routing objective.

Moreover, privacy constraints must align with evolving legal frameworks (e.g., GDPR~\cite{rossello2025llm}), which impose requirements on auditability, consent, and data minimization. Enforcing such constraints may require routable privacy policies or compliance-aware scoring functions that guide routing decisions alongside cost and performance metrics.

A promising research direction is the development of privacy-preserving routing algorithms that balance utility and confidentiality, e.g., by training scoring functions that jointly consider model performance and data sensitivity, or by incorporating real-time privacy budgets. Federated and decentralized training setups~\cite{kuang2024federatedscope} may further reduce privacy risks by avoiding centralized data storage.

In sum, integrating privacy and security as first-class constraints into multi-LLM inference systems is essential for deploying these technologies in sensitive domains. Future systems should aim to support explainable, enforceable privacy-aware routing that is compatible with real-world usage patterns and legal mandates~\cite{yuan2024fellas}.
\\
\\
The challenges discussed above reveal that designing multi-LLM inference systems implies satisfying multiple requirements across multiple dimensions, including modality integration, distributed coordination, system evaluation, and data privacy. To consolidate these directions, Table~\ref{tab:open_questions} summarizes a set of pressing open research questions categorized by their corresponding technical areas. While not exhaustive, this overview is meant to guide future investigations and highlight the inherently interdisciplinary nature of the problem space.

\section{Conclusion}\label{sec6}
This survey presents a comprehensive analysis of routing and HI techniques designed to optimize LLM inference under real-world constraints. These techniques demonstrate that routing and HI frameworks can substantially reduce resource consumption without significantly compromising performance, making them well-suited for budget-aware, latency-sensitive, and edge-centric deployments. By systematically evaluating recent approaches across key dimensions—including compute, memory, energy, latency, cost, scalability, and modality—we identify both the strengths and limitations shaping the current state of the field.

Yet, despite the gains achieved by routing and HI, their use in edge computing/intelligence systems is still in a nascent phase. Critical aspects such as modality support, large-scale scalability, and energy-awareness are either underdeveloped or entirely missing in most existing systems. Additionally, while many methods excel in optimizing specific constraints, few provide balanced, general-purpose solutions that can adapt across edge deployment settings. This suggests an urgent need for more holistic frameworks that integrate resource-awareness with the growing demands of multimodal and distributed applications.

Looking ahead, the future of efficient LLM inference lies in combining the adaptability of routing with the architectural ingenuity of model optimization strategies. As new LLMs continue to emerge with diverse capabilities and cost profiles, dynamic, context-sensitive inference pipelines will become increasingly essential. We hope this survey serves as both a reference point and a call to action—highlighting promising directions for future research while encouraging a broader shift toward smarter, more sustainable model deployment practices.

\bibliographystyle{ieeetran}
\bibliography{refs}

\end{document}